\documentclass[preprint,5p,times,twocolumn]{elsarticle}

\usepackage{amssymb}
\usepackage{amsmath}
\usepackage{svg}
\usepackage{bm}
\usepackage{booktabs}
\usepackage{comment}
\usepackage{adjustbox}
\usepackage{float}
\usepackage{gensymb}
\usepackage{fancyhdr}
\usepackage{url}
\usepackage[compact]{titlesec}
\usepackage[
 font=footnotesize % or small or scriptsize
 ]{subfig}
\usepackage{lineno}
\journal{Pattern Recognition Letters. Published version: \url{https://doi.org/10.1016/j.patrec.2025.02.016}}

\titlespacing{\section}{0.2pt}{0.4ex}{0.2ex}
\titlespacing{\subsection}{0.2pt}{0.4ex}{0.2ex}
\titlespacing{\subsubsection}{0pt}{0.2ex}{0.2ex}

\begin{document}

\begin{frontmatter}



\title{ConsistentDreamer: View-Consistent Meshes Through Balanced Multi-View Gaussian Optimization}

%%\author[label1,label2]{}

\author[a,b]{Onat Şahin}
\ead{onat.sahin@huawei.com, onat.sahin@tum.de}
\author[a]{Mohammad Altillawi\corref{cor1}}
\ead{mohammad.altillawi1@huawei.com}
\author[a]{George Eskandar}
\ead{george.basem.eskandar@huawei.com}
\author[a]{Carlos Carbone}
\ead{carlos.salvador.lorio@huawei.com}
\author[a]{Ziyuan Liu}
\ead{ziyuan.liu1@huawei.com}
\cortext[cor1]{Corresponding author.\\ \indent© 2025. This manuscript version is made available under the CC-BY-NC-ND 4.0 license https://creativecommons.org/licenses/by-nc-nd/4.0/}

\affiliation[a]{organization={Intelligent Cloud Technologies Laboratory, Huawei Munich Research Center},
            addressline={Riesstraße 25}, 
            city={Munich},
            postcode={80992}, 
            %state={},
            country={Germany}
            }
\affiliation[b]{organization={School of Computation, Information and Technology, Technical University of Munich},%Department and Organization
            addressline={Arcisstraße 21}, 
            city={Munich},
            postcode={80333}, 
            %state={},
            country={Germany}
            }
%Arcisstraße 21, Munich, 80333, Germany
\begin{abstract}
Recent advances in diffusion models have significantly improved 3D generation, enabling the use of assets generated from an image for embodied AI simulations. However, the one-to-many nature of the image-to-3D problem limits their use due to inconsistent content and quality across views. Previous models optimize a 3D model by sampling views from a view-conditioned diffusion prior, but diffusion models cannot guarantee view consistency. Instead, we present ConsistentDreamer, where we first generate a set of fixed multi-view prior images and sample random views between them with another diffusion model through a score distillation sampling (SDS) loss. Thereby, we limit the discrepancies between the views guided by the SDS loss and ensure a consistent rough shape. In each iteration, we also use our generated multi-view prior images for fine-detail reconstruction. To balance between the rough shape and the fine-detail optimizations, we introduce dynamic task-dependent weights based on homoscedastic uncertainty, updated automatically in each iteration. Additionally, we employ opacity, depth distortion, and normal alignment losses to refine the surface for mesh extraction. Our method ensures better view consistency and visual quality compared to the state-of-the-art. Project page: \url{https://onatsahin.github.io/ConsistentDreamer/}
\end{abstract}

\begin{comment}
\begin{graphicalabstract}
\centering
\includegraphics[scale=1]{teaser_image_no_dq_plain_lq_v3_pdf.pdf}
\end{graphicalabstract}

\begin{highlights}
\item We propose an image-to-3D Gaussian optimization method for better view consistency.
\item We guide unseen views with multi-view priors using SDS loss for consistent shape.
\item We obtain consistent fine details with multi-view pixel-level supervision.
\item We estimate task uncertainty through backpropagation to balance loss components.
\item We achieve better view consistency than state-of-the-art methods.
%\item Prompt selection from multi-view for SDS optimization increases view-consistency.
%\item Pixel-level supervision with multiple views increases texture quality in image-to-3D.
%\item Estimating task uncertainty through backpropagation helps balance loss components.
%\item Opacity, normal alignment losses improve mesh extraction from Gaussians.
\end{highlights}
\end{comment}

%% Keywords

\end{frontmatter}

%%\linenumbers


\section{Introduction}
\label{sec1}
%% Labels are used to cross-reference an item using \ref command.

Training and testing embodied AI agents in the real world is challenging due to environmental conditions, resource requirements, and safety concerns, making simulations central to robotics research \cite{gen2sim}. High-fidelity 3D assets are key to simulating tasks that require visual perception, such as grasping the robotic arm. However, the availability of 3D assets is limited, with even the largest datasets \cite{objaverseXL} being orders of magnitude smaller than image-text datasets. Traditionally, 3D assets are handcrafted or reconstructed through methods like NeRF \cite{mildenhall2020nerf} and 3D Gaussian Splatting \cite{kerbl3Dgaussians}. However, these approaches are time-consuming and labor-intensive. Advancements in AI have made 3D asset generation from a single image promising. However, image-to-3D remains under-constrained, with multiple plausible outputs for the same front view.

Recently, diffusion models have thrived in image generation from text prompts, demonstrating strong generalization abilities. This success has inspired their utilization for 3D asset generation. The seminal work DreamFusion \cite{poole2022dreamfusion} adopts a 2D diffusion model to iteratively enhance random views in a NeRF representation, obtaining a 3D representation from a text prompt. However, this approach struggles with consistent view generation due to the lack of a 3D prior, leading to mismatched views known as the Janus problem. A potential solution is to include 3D priors in the pipeline, such as diffusion models trained on novel view synthesis. For example, Zero-1-to-3 \cite{Liu_2023_ICCV} generates a complementary view of an object given one input view and a relative viewpoint, influencing the development of other view-conditioned generation methods for image-to-3D tasks.

There are three main approaches to using view-conditioned diffusion models for image-to-3D. The first follows DreamFusion's iterative approach but uses a view-conditioned model as a 3D prior instead of a text-conditioned 2D prior. Models like \cite{tang2024dreamgaussian, qian2024magic} use Zero-1-to-3-based models for this purpose, enabling consistent base shape generation and avoiding the Janus problem. Gen2Sim \cite{gen2sim} applies this method to automate asset generation for robotic skill reinforcement learning. However, this approach struggles with finer details, leading to a lack of detail and inconsistent quality in views farther from the input image. 

The second approach uses diffusion models to generate multiple views in the first stage, then fits a 3D parametric model like NeuS \cite{wang2021neus} with a reconstruction loss \cite{Liu_2023_ICCV, shi2023MVDream, wang2023imagedream, shi2023zero123plus, liu2024syncdreamer, Woo_2024_CVPR, Long_2024_CVPR}. Generating multiple prior views at once improves consistency compared to iterative random view generation, but results in long optimization times due to the parametric models and unrefined geometry and texture due to limited view information.

The third approach trains specialized feed-forward models that use multi-view images as input to directly predict 3D representations \cite{hong2024lrm, tang2024lgm, wang2024crm, tochilkin2024triposr, Zou_2024_CVPR}. This method avoids the bottlenecks of the previous approach and achieves the fastest generation speeds. However, the encoder-decoder structure leads to quality degradation from the input images, causing blurry textures and inconsistent quality in the output.
\begin{figure}[tpb]
\includegraphics[scale=0.62]{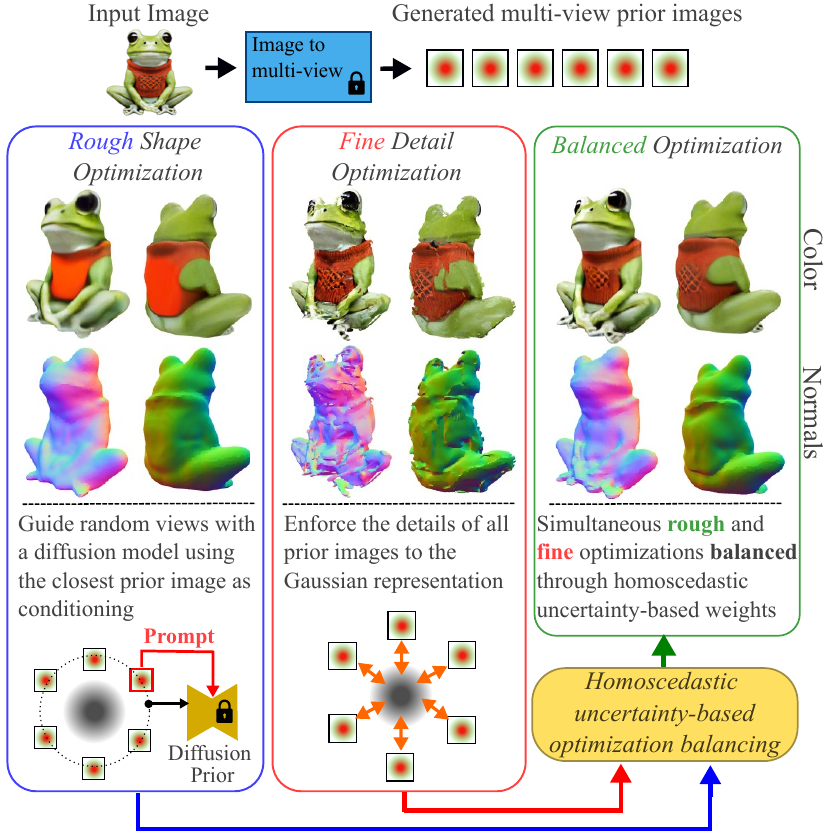}
\caption{\textbf{ConsistentDreamer for image-to-3D.} Unlike prior optimization methods using only view-conditioned diffusion guidance or multi-view reconstruction, we utilize both for balanced optimization of rough shape and fine details, improving view consistency in content and visual quality.}
\label{fig:teaser}
\end{figure}

We propose ConsistentDreamer (Fig. \ref{fig:teaser}), an optimization-based image-to-3D method that generates view-consistent 3D shapes with finer details across all views. In an initial stage, we use a multi-view diffusion prior to generate consistent multi-view images from a single image, which serve as references during optimization. In each optimization step, a second diffusion prior guides the generation of a random novel view. Unlike previous methods that use the same single input image as the condition for all views, we select the generated prior image closest to the guided view. By ensuring the guided random view remains close to the condition image, we solve the inconsistencies seen in view-conditioned optimization.

This optimization gives us a consistent but rough base shape. To optimize the finer details, we exploit the iterative nature of our solution and simultaneously enforce the details of the generated prior view images in every step. This fine detail optimization directly compares our view images against corresponding renders of our representation at the pixel level. Keeping the supervision at the pixel level allows us to avoid the problems of degraded color and detail. 

By using generated multi-view prior images to optimize fine details and guide the generation of unseen views for rough shape optimization simultaneously, we achieve a more complete coverage of views around the object compared to previous methods. This results in representations with improved view consistency in both geometry and visual quality.

The rough base shape and fine detail optimizations operate in embedded and image spaces, respectively, leading to a difference in loss magnitudes. This leads to an imbalance when using them simultaneously. To solve this, we introduce task-dependent weights to these loss components based on homoscedastic uncertainty. These weights are updated automatically in each iteration based on our loss, ensuring a balanced optimization with no additional supervision or handpicked loss schedules. We use a Gaussian representation due to its advantages in rendering and optimization speeds, and our balanced optimization results in final representations with smaller Gaussian counts that contain fewer redundancies.

For compatibility with embodied AI simulations, we pick 3D meshes as our primary output despite the Gaussian representation being unfit for mesh extraction. To alleviate this, we add opacity, normal alignment, and depth distortion losses to our optimization to align the Gaussians better with the object surface.

In summary, our contributions are: (1) We introduce ConsistentDreamer, an image-to-3D optimization pipeline that leverages generated multi-view prior images to optimize fine details while simultaneously utilizing them to guide the generation of unseen views. (2) We propose to find an optimal balance between rough base shape and fine detail optimizations utilizing task-dependent dynamic weights that are automatically updated based on homoscedastic uncertainty. (3) We address Gaussians’ shortcomings in surface representation using opacity and depth-based alignment losses within an iterative image-to-3D pipeline. (4) Our method ensures a more comprehensive view coverage compared to previous methods, resulting in improved view consistency and visual quality. 

%% Use \subsection commands to start a subsection.
\section{Related Work} \label{related_work}
Recent developments in 3D representations have improved generating 3D assets from reference images. Some works modified NeRFs for various use cases \cite{mildenhall2020nerf, Yu_2021_CVPR, 9878785, trevithick2021grf, ZIMNY20248, JIN2024160}, but the major breakthroughs were driven by diffusion models.

\noindent \textbf{SDS-based image-to-3D. }DreamFusion \cite{poole2022dreamfusion} introduced the score distillation sampling (SDS) loss, using text-to-image diffusion models iteratively to optimize a NeRF representation. This approach has been adapted to Image-to-3D in various ways \cite{Tang_2023_ICCV, qian2024magic}, with DreamGaussian \cite{tang2024dreamgaussian} optimizing a Gaussian representation \cite{kerbl3Dgaussians} to improve optimization times. In our method, we optimize a Gaussian representation iteratively while increasing the efficiency through loss balancing with weights based on homoscedastic uncertainty. We also improve the geometry and texture consistency with multi-view image references. 

\noindent \textbf{Multi-view reconstruction-based image-to-3D. }To move away from instance-based optimization and achieve fast inference and generalizability, many methods train large models using 3D datasets such as Objaverse XL \cite{objaverseXL}. Some such methods focus on consistent multi-view image generation \cite{Liu_2023_ICCV, shi2023MVDream, wang2023imagedream, shi2023zero123plus, liu2024syncdreamer, Woo_2024_CVPR, Long_2024_CVPR} and rely on learning-based reconstruction methods like NeuS \cite{wang2021neus} for 3D reconstruction. However, NeuS bottlenecks these methods in inference time and the success of transferring the image quality to 3D. In our method, we use multi-view images as references within an SDS-based iterative optimization to form the shape of the Gaussians. Therefore, our method does not rely on external reconstruction methods, and does not depend solely on the reference views due to the use of a diffusion prior to optimize the rough base shape.% with an SDS loss.

\noindent \textbf{Feed-forward image-to-3D. } Some methods predict 3D representations directly from one or more images end-to-end by training large models on 3D datasets \cite{hong2024lrm, tang2024lgm, wang2024crm, tochilkin2024triposr, Zou_2024_CVPR}. While these methods offer the fastest inference times and the highest output quality among off-the-shelf methods, the assets generally suffer from imperfect geometry and texture, especially in non-frontal views. Another shortcoming of these methods is blurry textures due to the feed-forward encoder-decoder structure diminishing the finer details of the input images. We largely avoid this problem with our fine detail optimization, supervising our Gaussian representation with multi-view prior images at the image level at every step of our optimization instead of using them in a feed-forward manner.
%\begin{comment}
\section{Preliminary} \label{preliminary}

\noindent \textbf{3D Gaussian Splatting. }3DGS \cite{kerbl3Dgaussians} is a 3D representation in which a scene is comprised of a set of 3D Gaussians \(\{G_i\}_{i=1}^M\), where the \(i\)-th Gaussian is defined as:
\begin{equation}
        G_i = \{\bm{\mu}_i, \bm{q}_i, \bm{s}_i, \bm{c}_i, \alpha_i\}.
\end{equation}
Here, \(\bm{\mu}_i \in \mathbb{R}^3\) is the center of the Gaussian, \(\bm{s}_i \in \mathbb{R}^3\) is the scaling factor, \(\bm{q}_i \in SO(3)\) is the rotation quaternion, \(\bm{c}_i \in \mathbb{R}^3\) is the color, and \(\alpha_i \in \mathbb{R}\) is the opacity. The process of 3DGS involves rasterizing these 3D Gaussians \(\{G_i\}_{i=1}^M\) by sorting them in depth order in the camera space and projecting them onto the image plane. The rasterization operation is differentiable, enabling iterative optimization with stochastic gradient descent. \\
\noindent\textbf{Score Distillation Sampling (SDS) Loss. }
First proposed by DreamFusion \cite{poole2022dreamfusion}, score distillation sampling (SDS) loss is a differentiable loss function used to iteratively optimize a 3D representation by leveraging a text-to-image or image-to-image diffusion prior to 3D. During optimization, in each iteration, a viewpoint $p$ around the optimized 3D representation $\Theta$ is sampled for rasterization. Random Gaussian noise $\epsilon$ is added to the rasterized image $I^p_\text{RGB}$ and passed to the diffusion model $\phi$. In the image-to-3D case, the diffusion model is prompted with an input image $\tilde{I}^r_\text{RGB}$ and the elevation/azimuth differences between the input viewpoint and the sampled viewpoint $\Delta p$. The diffusion model predicts the added noise $\epsilon_\phi(.)$. This prediction is evaluated against $\epsilon$ and weighted with the weighting function $w(t)$ based on timestep $t$, forming score distillation sampling (SDS) loss, formulated by DreamGaussian \cite{tang2024dreamgaussian} as:
\begin{equation}
\label{eqn:base_sds}
    \nabla_\Theta\mathcal{L}_\text{SDS} = \mathbb{E}_{t,p,\epsilon} \left[ w(t)(\epsilon_\phi(I^p_\text{RGB};t, \tilde{I}^r_\text{RGB}, \Delta p) - \epsilon)\frac{\partial I^p_\text{RGB}}{\partial\Theta} \right].
\end{equation}
%\end{comment}
\begin{figure*}[h!]
      \centering
      \includegraphics[scale=0.8]{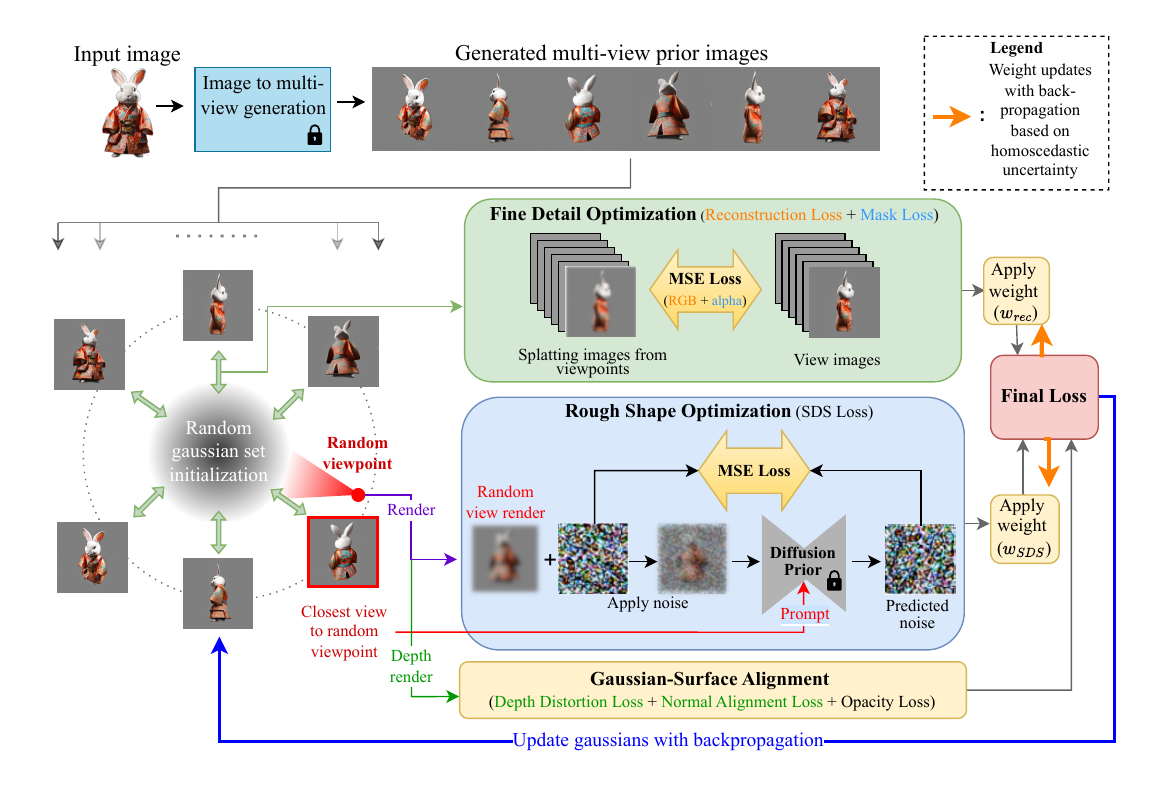}
      \caption{\textbf{ConsistentDreamer pipeline.} ConsistentDreamer is a Gaussian-based method for view-consistent 3D generation from a single image, guided by consistent multi-view images generated in a prior stage. Rough shape is optimized by improving random views with a diffusion conditioned on the closest prior view, while fine details are refined by comparing all prior views to corresponding views of the representation. A balance between rough and fine optimizations is found with dynamic weights updated based on the final loss, with mesh extraction ensured through depth distortion, normal alignment, and opacity losses.}
    \label{method-overview} 
\end{figure*}
\section{Method}
In our method, we iteratively optimize a Gaussian representation for the Image-to-3D task, as shown in Fig. \ref{method-overview}. First, we generate a set of $N=6$ multi-view images with fixed viewpoints $\{r_{i}\}^N_{i=1}$ from the input image using a multi-view generation model. In each iteration, we use these prior view images in two ways: as prompts for a diffusion prior to calculate the SDS loss, optimizing the rough shape, and as ground truths for a reconstruction loss, comparing the prior views to corresponding renders to refine finer details. To balance the rough shape and fine detail optimizations that operate on different loss scales and domains, we introduce dynamic loss weights learned through homoscedastic uncertainty. Additionally, we apply an opacity loss that removes semi-transparent Gaussians, a depth distortion loss that aligns Gaussians with the surface, and a normal alignment loss that optimizes Gaussian orientations to ensure accurate surface representation for mesh extraction. \\
\textbf{Rough Shape Optimization. }For rough shape optimization, we use an SDS-based optimization previously described in section \ref{preliminary} with a pre-trained Zero-1-to-3 model \cite{Liu_2023_ICCV} as the diffusion prior. This model generates novel views based on an input image and relative elevation and azimuth angles. We observed that the consistency between generated views drops as the angles between the input and the generated view increase, leading to quality and content inconsistencies across views in the final representation. We avoid this problem by integrating our generated prior view images into the optimization. Instead of using the same image as the prompt in each iteration, we pick the generated view $\Tilde{I}^{r_i}$ closest to the sampled rasterized view $I^p$ as the prompt image, based on the angular distance between their viewpoints $r_i$ and $p$, where:
\begin{equation}
    \Delta p_i = (\theta_p - \theta_{r_i}, \beta_p - \beta_{r_i})
\end{equation}\begin{flalign}
    \begin{split}
    |\Delta p_i| &= |arccos(cos(\beta_p + \frac{\pi}{2})cos(\beta_{r_i} + \frac{\pi}{2}) \\
    &+ sin(\beta_p + \frac{\pi}{2})sin(\beta_{r_i} + \frac{\pi}{2})cos(\theta_p - \theta_{r_i}))|
    \end{split}
\end{flalign}
$\Delta p_i$ denotes the relative azimuth ($\theta$) and elevation ($\beta$) angles between viewpoints $p$ and $r_i$, and $|\Delta p_i|$ denotes angular distance. Therefore, our SDS loss is formulated as:
\begin{equation}
    \nabla_\Theta\mathcal{L}_\text{SDS} = \mathbb{E}_{t,p,\epsilon} \left[ w(t)(\epsilon_\phi(I^p_\text{RGB};t, \Tilde{I}^{r_{i^*}}_{\text{RGB}}, \Delta p) - \epsilon)\frac{\partial I^p_\text{RGB}}{\partial\Theta} \right]
\end{equation}
where $i^*={argmin}_i |\Delta p_i|$. Picking the closest prompt image among six view images ensures that $\Delta p_i$ remains small, leading to more consistent predictions from our diffusion prior. This way, we are able to optimize a base shape that is consistent across all views. \\ 
\textbf{Fine Detail Optimization. }Concurrent to the rough shape optimization with the SDS loss, we also use the generated prior views as ground truth to calculate mean square error and obtain reconstruction and mask losses in each iteration: 
\begin{equation}
    \mathcal{L}_{\text{reconstruction}} = \frac{1}{N}\sum^N_i \lambda_{\text{RGB}} \Vert I^{r_i}_{\text{RGB}} - \tilde{I}^{r_i}_{\text{RGB}} \Vert^2_2
\end{equation}
\begin{equation}
    \mathcal{L}_{\text{mask}} = \frac{1}{N}\sum^N_i \lambda_{\text{A}}\Vert I^{r_i}_{\text{A}} - \tilde{I}^{r_i}_{\text{A}} \Vert^2_2
\end{equation}
The reconstruction loss enforces the fine details of the images to the representation, while the mask loss eliminates unwanted floaters. Here, $N$ denotes the number of generated views. $\tilde{I}^{r_i}_{\text{RGB}}$ is the $i$th reference image, while $I^{r_i}_{\text{RGB}}$ is the rasterization of the Gaussian from the corresponding viewpoint. $\tilde{I}^{r_i}_{\text{A}}$ and $I^{r_i}_{\text{A}}$ are the alpha channels of $\tilde{I}^{r_i}_{\text{RGB}}$ and $I^{r_i}_{\text{RGB}}$ respectively. $\lambda_{\text{RGB}}$ and $\lambda_{\text{A}}$ are fixed coefficients. \\%For $\mathcal{L}_{\text{rec}}$ we use mean square error. \\ 
\textbf{Opacity Loss. }With multi-view SDS, reconstruction, and mask losses, we obtain a Gaussian representation with improved detail and consistency all around the object. However, Gaussian representations are not readily suitable for mesh extraction, as the object surface is not clearly defined due to many semi-transparent Gaussians. Motivated by \cite{Chen2023NeuSGNI, Guedon_2024_CVPR}, we drive the opacities of the Gaussians (denoted by $\alpha_k$ for $k$-th Gaussian) to 1 (fully opaque), using
\begin{equation}
    \begin{aligned}
        \mathcal{L}_{\text{opacity}} = \lambda_{\text{opacity}} \frac{1}{N}\sum^N_k -\alpha_k log(\alpha_k + 10^{-10}) \\
        - (1-\alpha_k)log(1 - \alpha_k + 10^{-10})
    \end{aligned}
\end{equation}
\indent This way, we eliminate semi-transparent Gaussians during optimization to achieve better-defined, more complete surfaces during mesh extraction. \\ 
\textbf{Depth Distortion and Normal Alignment Losses. }Despite achieving a representation with opaque Gaussians and consistent detail all around the object, finer surface details of the object still cannot be captured during mesh extraction. To alleviate this, we include depth-normal alignment ($\mathcal{L}_{\text{normal}}$) and depth distortion ($\mathcal{L}_{\text{depth}}$) losses following \cite{zhang2024radegsrasterizingdepthgaussian}. After rasterizing Gaussians from the randomly sampled viewpoint during each step, the depth distortion loss brings Gaussians within each ray used during rasterization closer for a better-defined surface. The depth-normal alignment adjusts the Gaussians so that their normals align with normals approximated from rasterized depth maps, allowing for more accurate and less noisy surfaces. Refer to \cite{zhang2024radegsrasterizingdepthgaussian} for the details on the rasterizer and the loss formulations. \\ 
\textbf{Balancing Rough and Fine Optimizations.} In our initial experiments, we observed that the optimization of rough shape and fine details is unstable, with fluctuations in the corresponding losses leading to unnecessary densification of the Gaussian representation. Despite both tasks using photometric error (mean square error), they require different optimization schedules due to the scale differences in their losses, which arise from operating in different spaces. Specifically, the SDS loss for rough shape optimization is evaluated in an embedded space, while the reconstruction loss for fine detail optimization is evaluated in image space. To optimize the schedule for both loss components, we introduce dynamic loss weights based on homoscedastic uncertainty, a task-dependent uncertainty that remains constant across inputs. In multi-task settings, this uncertainty captures task confidence, allowing for proper weighting. \cite{8578879}. We optimize the weights $\omega_{\text{SDS}}$ and $\omega_{\text{rec}}$, estimating homoscedastic uncertainties of our SDS and reconstruction losses respectively, using backpropagation with respect to the full loss function $\partial\mathcal{L}_{\text{full}} / \partial w $ using an Adam optimizer. We apply the weights and calculate $\mathcal{L}_{\text{rough}}$ and $\mathcal{L}_{\text{fine}}$ as
\begin{equation}
        \mathcal{L}_{\text{rough}} = e^{-\omega_{\text{SDS}}}\mathcal{L}_{\text{SDS}} +\omega_{\text{SDS}}
\end{equation}
\begin{equation}
        \mathcal{L}_{\text{fine}} = e^{-\omega_{\text{rec}}}\mathcal{L}_{\text{reconstruction}} +\omega_{\text{rec}} + \mathcal{L}_{\text{mask}}
\end{equation}
The weights $\omega_{\text{SDS}}$ and $\omega_{\text{rec}}$ are independent scalar values where $\omega := \log\sigma^2$ with $\sigma^2$ being the actual uncertainty, not directly used due to numerical stability. The weights are initialized as 0, defaulting to a non-weighted loss. As these weights increase over the course of the optimization, the first term $e^{-\omega}\mathcal{L}$ tempers the loss, while the second term $+\omega$ acts as regularization to prevent infinite uncertainty. With this weighting, we achieve an optimization with more stable loss values, resulting in more efficient representations with less number of Gaussians.  
\newline
\indent Adding all these together, the final loss becomes:
\begin{equation}
    \begin{aligned}
        \mathcal{L}_{\text{full}} = \mathcal{L}_{\text{rough}} + \mathcal{L}_{\text{fine}} + \mathcal{L}_{\text{opacity}} + \mathcal{L}_{\text{normal}} + \mathcal{L}_{\text{depth}}
    \end{aligned}
\end{equation}

\section{Experiments}
\begin{figure*}[th!]
      \centering
      \includegraphics[scale=0.75]{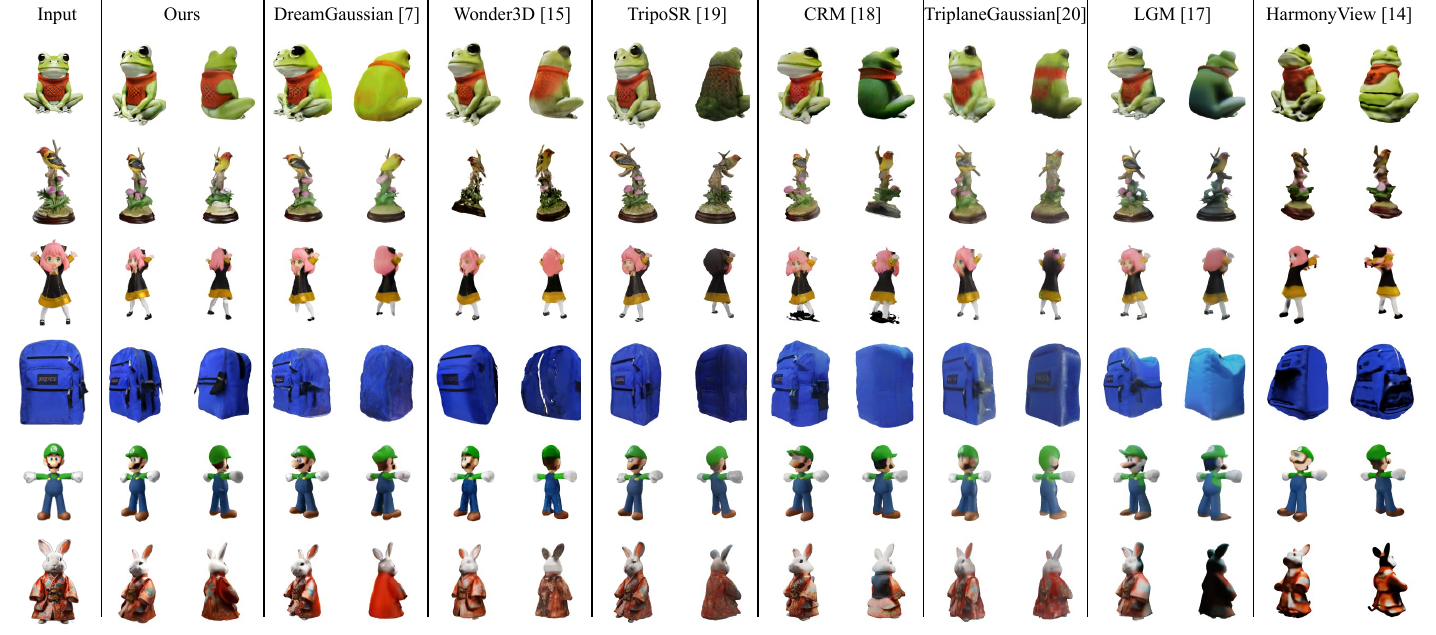}
      \caption{\textbf{Qualitative comparisons on image-to-3D mesh generation.} We compare ConsistentDreamer against various methods using images from the internet. We include the input image and two diagonal views of each generated mesh from 45$\degree$ and 225$\degree$ azimuth angles. Our method gives the best results with consistent detail and color across views all around the object.}
      \label{fig:qualitative}
\end{figure*}

\begin{figure*}[th!]
      \centering
      \includegraphics[scale=0.75]{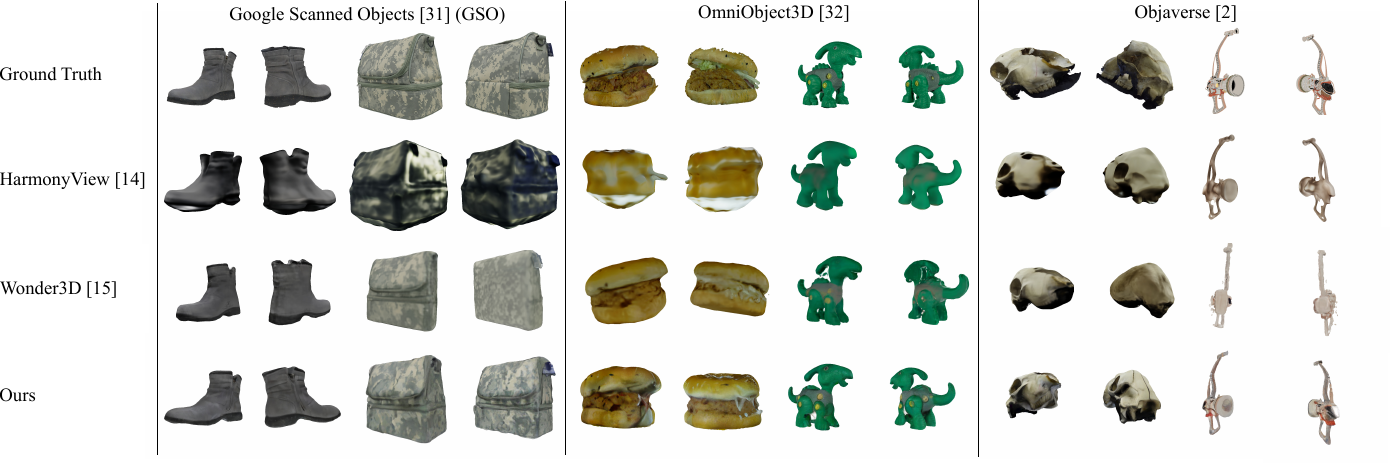}
      \caption{\textbf{Sample Meshes from our quantitative evaluation.} We show some sample meshes from our quantitative evaluation shown on Tables \ref{quantitive_table_gso}, \ref{quantitive_table_oo} and \ref{quantitive_table_obja}. We include the two diagonal views of each generated mesh from 45$\degree$ and 225$\degree$ azimuth angles, along with corresponding ground truth views.}
      \label{fig:qualitative_dataset}
\end{figure*}

\subsection{Experimental Setup}
\noindent\textbf{Implementation Details.} We optimize the 3D Gaussians for 500 iterations. We generate six prior view images from fixed viewpoints defined by elevation and azimuth angles $(20\degree, 30\degree)$, $(-10\degree, 90\degree)$, $(20\degree, 150\degree)$, $(-10\degree, 210\degree)$, $(20\degree, 270\degree)$ and $(-10\degree, 330\degree)$ using Zero123++ \cite{shi2023zero123plus}. For pre-processing these images, initializing and optimizing the Gaussians, and sampling camera poses, we rely on parameters specified by DreamGaussian \cite{tang2024dreamgaussian}. We set the camera's y-axis FoV to 30\degree. To ensure consistency, we only use the generated images and not the original input image during optimization. We set fixed loss coefficients as $\lambda_{\text{SDS}}=1$, $\lambda_{\text{RGB}}=10000$, $\lambda_{\text{A}}=1000$, $\lambda_{\text{opacity}}=0.1$, $\lambda_{\text{normal}}=0.05$, $\lambda_{\text{depth}}=100$, based on previous work. We extract meshes and apply texture enhancement for 50 steps by the extending the method proposed by DreamGaussian for multi-view image references, using the generated images from $(20\degree, 30\degree)$, $(20\degree, 150\degree)$, $(20\degree, 270\degree)$ viewpoints for direct reconstruction. All the experiments are run on a single GPU with 24 GB memory, with our method generating a textured mesh from a single image in about a minute.\\
\textbf{Baselines.} We evaluate our method against various of state-of-the-art image-to-3D generation works that cover the approaches summarized in section \ref{related_work}: DreamGaussian \cite{tang2024dreamgaussian}, LGM \cite{tang2024lgm}, HarmonyView \cite{Woo_2024_CVPR}, Wonder3D \cite{Long_2024_CVPR}, TriplaneGaussian \cite{Zou_2024_CVPR}, CRM \cite{wang2024crm}, and TripoSR \cite{tochilkin2024triposr}. \\
\textbf{Datasets.} We evaluate our method quantitatively on samples from Google Scanned Objects (GSO) \cite{Downs2022GoogleSO}, OmniObject3D \cite{wu2023omniobject3d} and Objaverse \cite{objaverseXL} datasets. GSO is comprised of real scanned common household items. OmniObject3D is a real scanned objects dataset with the highest number of classes. In contrast, Objaverse is the largest available 3D dataset collected from the internet, comprised of both real object scans and digital assets created with industry standard software. For GSO experiments, we use the evaluation set comprised of 30 objects compiled by \cite{liu2024syncdreamer}. For OmniObject3D and Objaverse experiments, we randomly select 30 objects from different classes. For each object, we render 8 views with zero elevation, separated by 45$\degree$ as ground truth, and use the front view as input for 3D generation. Additionally, we perform qualitative evaluation on front-view images with zero elevation selected from the internet. \\ 
\textbf{Metrics.} For quantitative evaluation, we obtain 8 side renders of the reconstructed meshes, separated by 45\degree, matching the ground truth mesh renders. We evaluate SSIM, PSNR, LPIPS, and CLIP similarity between corresponding renders. Additionally, we assess CLIP consistency by calculating the mean CLIP similarity between the input image and each of the 8 renders from the generated mesh, measuring the overall consistency of the generated mesh with the input image.
\subsection{Comparisons with State-of-the-Art}
\begin{figure}[t!]
      \centering
      \adjustbox{trim=0.8cm 0cm 0cm 0cm, clip=true}{%
        \includegraphics[scale=1.3]{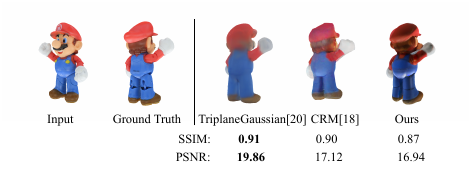}%
      }
      \caption{\textbf{SSIM and PSNR results.} To analyze the behavior of SSIM and PSNR we show back-views of generated meshes, along with the front-view input image and the ground truth back-view. The results show that these metrics may not reflect the similarity in detail level and clarity.}
      \label{fig:ssim_psnr}
\end{figure}

\noindent\textbf{Qualitative Evaluation.} Our qualitative results on internet images are shown in Fig. \ref{fig:qualitative}. While front-view results are generally consistent across all methods, state-of-the-art methods struggle with other views, showing inconsistent geometry and/or texture quality. Textures are especially problematic, with details such as the sweater of the frog example being inconsistent across views and color tones being generally degraded and darker in the back views. Our method outperforms the state-of-the-art by maintaining consistent detail and color all around the object. \\
\textbf{Quantitative Evaluation.} Our results on the GSO, OmniObject3D and Objaverse sets are shown in Tables \ref{quantitive_table_gso}, \ref{quantitive_table_oo} and \ref{quantitive_table_obja} respectively. The results place our method at the top of the state-of-the-art in LPIPS, CLIP similarity, and CLIP consistency, while achieving comparable performance in SSIM and PSNR. We note that SSIM and PSNR may not capture similarity in detail level and sharpness as well as learning-based metrics. This is illustrated in Fig. \ref{fig:ssim_psnr}, where SSIM and PSNR favor blurry, less detailed images over sharper and more detailed ones. In contrast, LPIPS and CLIP similarity results show that our method achieves high similarity with the ground truth, and CLIP consistency results show that our generated meshes are highly consistent with the input image. In Figure \ref{fig:qualitative_dataset}, we show some sample meshes from these experiments.
\begin{table}[tb!]
\normalsize
\caption{Quantitative evaluation of our method on Google Scanned Objects \cite{Downs2022GoogleSO} dataset}
\label{quantitive_table_gso}
\begin{center}
\resizebox{\columnwidth}{!}{
    \begin{tabular}{@{}lccccc@{}}\toprule
    Method & SSIM $\uparrow$ & PSNR $\uparrow$ & LPIPS $\downarrow$ & CLIP Sim. $\uparrow$ & CLIP Cons. $\uparrow$ \\
    
    \cmidrule{1-6} 
    
    DreamGaussian \cite{tang2024dreamgaussian} & 0.88 & 19.26 & 0.14 & 0.88 & 0.85\\
    HarmonyView \cite{Woo_2024_CVPR} & 0.87 & 16.32 & 0.16 & 0.82 & 0.78\\
    LGM \cite{tang2024lgm} & 0.88 & 17.94 & 0.16 & 0.86 & 0.83\\
    Wonder3D \cite{Long_2024_CVPR} & \textbf{0.89} & 18.74 & 0.14 & 0.89 & 0.85\\
    TriplaneGaussian \cite{Zou_2024_CVPR} & \textbf{0.89} & \textbf{19.43} & 0.14 & 0.85 & 0.85\\
    CRM \cite{wang2024crm} & \textbf{0.89} & 19.14 & 0.14 & \textbf{0.90} & \textbf{0.87}\\
    TripoSR \cite{tochilkin2024triposr} & 0.88 & 18.06 & 0.14 & 0.88 & 0.85\\
    Ours & 0.88 & 19.06 & \textbf{0.13} & \textbf{0.90} & \textbf{0.87}\\    
    \bottomrule
    \end{tabular}
}
\end{center}
\end{table}
\begin{table}[tb!]
\normalsize
\caption{Quantitative evaluation of our method on OmniObject3D \cite{wu2023omniobject3d} dataset}
\label{quantitive_table_oo}
\begin{center}
\resizebox{\columnwidth}{!}{
    \begin{tabular}{@{}lccccc@{}}\toprule
    Method & SSIM $\uparrow$ & PSNR $\uparrow$ & LPIPS $\downarrow$ & CLIP Sim. $\uparrow$ & CLIP Cons. $\uparrow$ \\
    
    \cmidrule{1-6} 
    
    DreamGaussian \cite{tang2024dreamgaussian} & 0.88 & 16.58 & 0.15 & 0.88 & 0.86\\
    HarmonyView \cite{Woo_2024_CVPR} & 0.87 & 15.72 & 0.15 & 0.83 & 0.80\\
    LGM \cite{tang2024lgm} & 0.89 & 16.60 & 0.15 & 0.88 & 0.86\\
    Wonder3D \cite{Long_2024_CVPR} & 0.89 & 16.81 & 0.14 & 0.88 & 0.87\\
    TriplaneGaussian \cite{Zou_2024_CVPR} & \textbf{0.90} & \textbf{18.01} & \textbf{0.13} & 0.87 & 0.85\\
    CRM \cite{wang2024crm} & 0.89 & 17.20 & 0.14 & 0.90 & 0.89\\
    TripoSR \cite{tochilkin2024triposr} & 0.89 & 17.01 & 0.14 & 0.90 & 0.89\\
    Ours & 0.89 & 17.80 & \textbf{0.13} & \textbf{0.91} & \textbf{0.90}\\
    
    \bottomrule
    \end{tabular}
}
\end{center}
\end{table}
\begin{table}[tb!]
\normalsize
\caption{Quantitative evaluation of our method on Objaverse \cite{objaverseXL} dataset. Since all the baselines contain a component trained or fine-tuned on Objaverse, for a fair comparison we only compare against the baselines that are similar to our own method, where the 3D shape is obtained with iterative optimization supervised by multi-view images generated with a model fine-tuned on Objaverse.}
\label{quantitive_table_obja}
\begin{center}
\resizebox{\columnwidth}{!}{
    \begin{tabular}{@{}lccccc@{}}\toprule
    Method & SSIM $\uparrow$ & PSNR $\uparrow$ & LPIPS $\downarrow$ & CLIP Sim. $\uparrow$ & CLIP Cons. $\uparrow$ \\
    
    \cmidrule{1-6} 
    
    DreamGaussian \cite{tang2024dreamgaussian} & \textbf{0.87} & \textbf{15.71} & 0.16 & 0.84 & 0.83\\
    HarmonyView \cite{Woo_2024_CVPR} & 0.86 & 14.56 & \textbf{0.15} & 0.83 & 0.82\\
    Wonder3D \cite{Long_2024_CVPR} & 0.87 & 15.10 & \textbf{0.15} & 0.84 & 0.84\\
    Ours & 0.86 & 15.52 & \textbf{0.15} & \textbf{0.86} & \textbf{0.85}\\
    
    \bottomrule
    \end{tabular}
}
\end{center}
\end{table}
\subsection{Ablation Study} 
\subsubsection{Loss Components}
\begin{figure}[t!]%
\centering
\subfloat[][Input]{\includegraphics[trim={1.2cm 1.2cm 1.2cm 1.2cm}, clip, width=.09\textwidth]{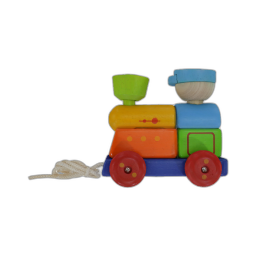}}%
\qquad
\subfloat[][Full opt.]{\includegraphics[trim={7cm 7cm 7cm 7cm}, clip, width=.09\textwidth]{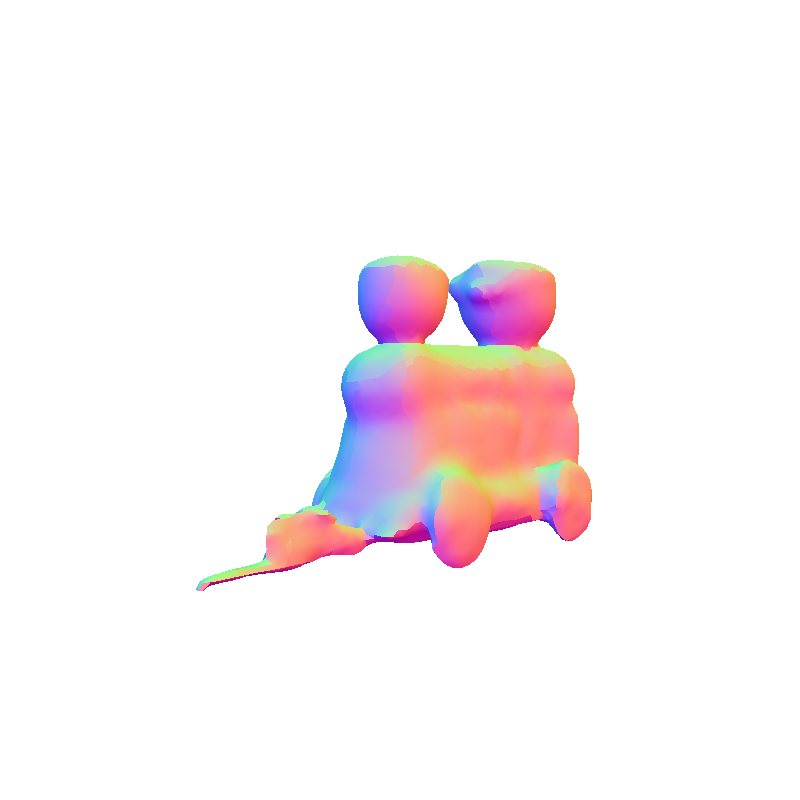}}%
\qquad
\subfloat[][w/o multi-view images]{\includegraphics[trim={7cm 7cm 7cm 7cm}, clip, width=.09\textwidth]{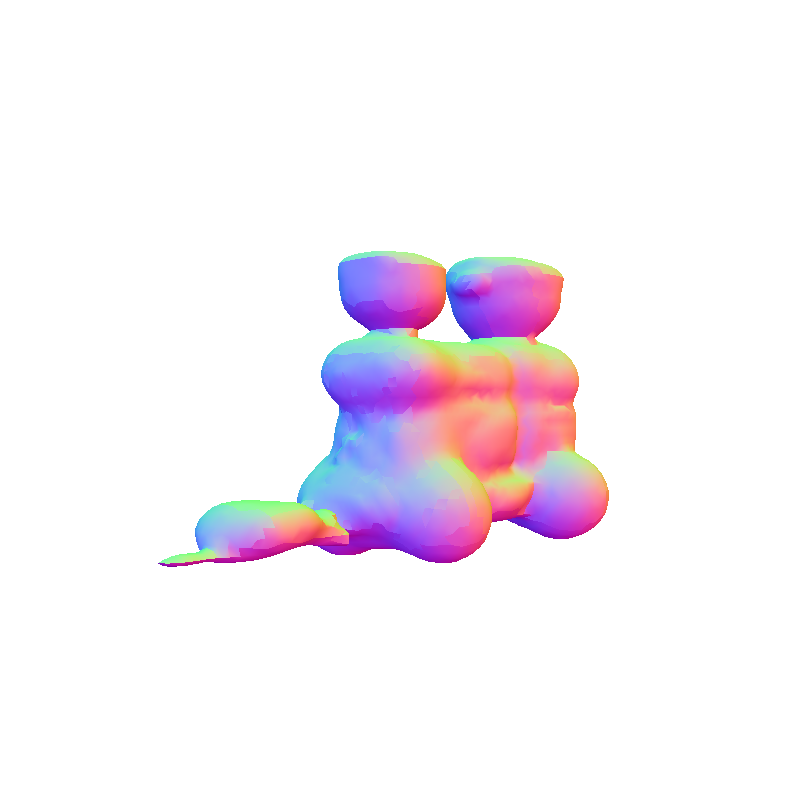}}%
\qquad
\subfloat[][w/o opt. balancing]{\includegraphics[trim={6cm 7cm 7cm 7cm}, clip, width=.09\textwidth]{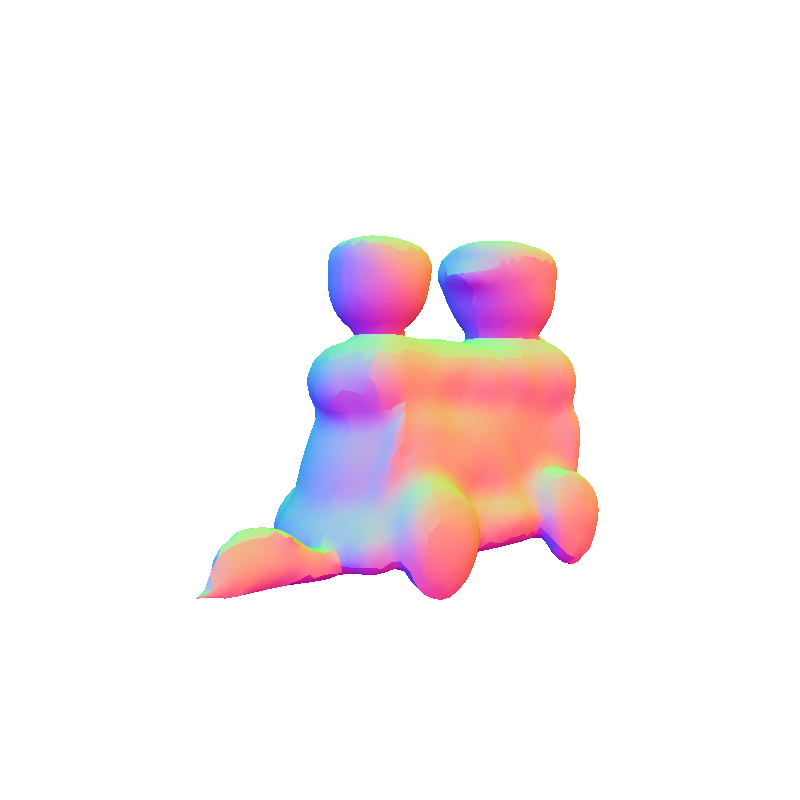}} \\
\subfloat[][w/o rough opt.]{\includegraphics[trim={6cm 7cm 7cm 7cm}, clip, width=.09\textwidth]{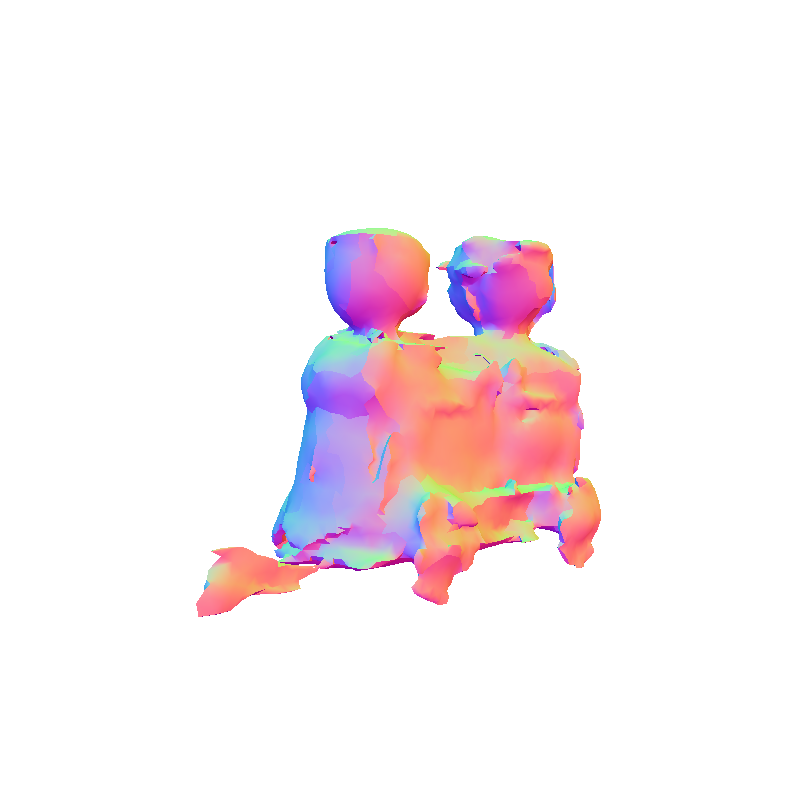}}
\qquad
\subfloat[][w/o fine opt.]{\includegraphics[trim={6cm 3cm 7cm 3cm}, clip, width=.06\textwidth]{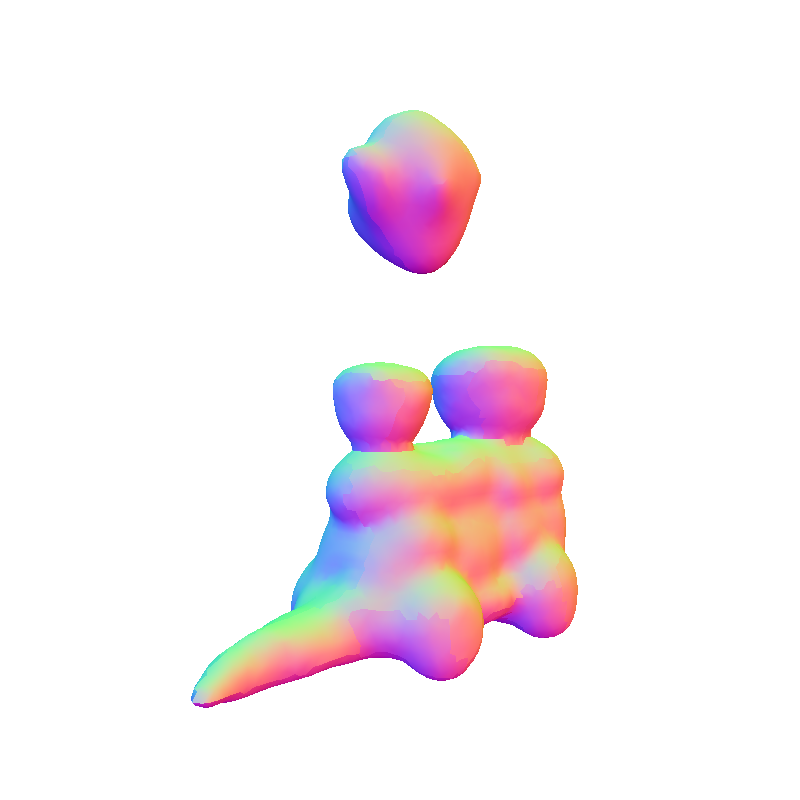}}
\qquad
\subfloat[][w/o opacity loss]{\includegraphics[trim={7cm 7cm 7cm 7cm}, clip, width=.09\textwidth]{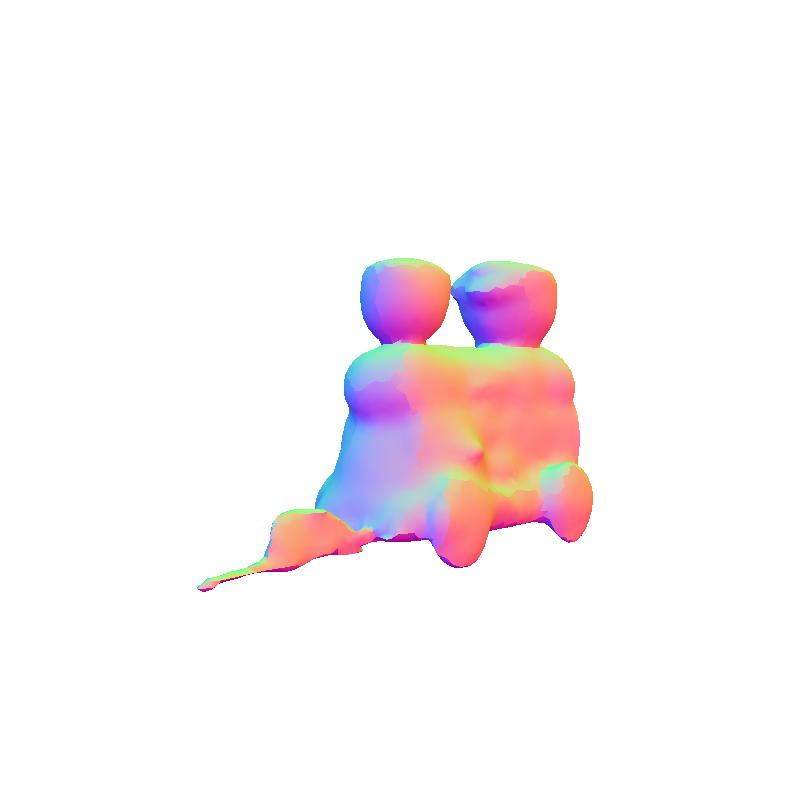}}
\qquad
\subfloat[][w/o depth \& normal loss]{\includegraphics[trim={7cm 7cm 7cm 7cm}, clip, width=.09\textwidth]{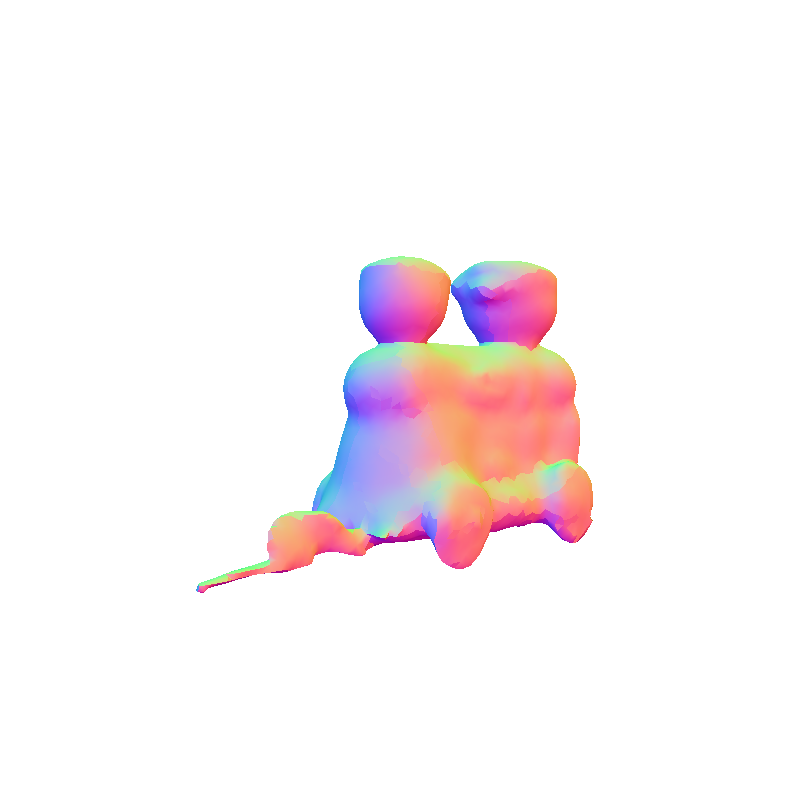}}
\caption{\textbf{Ablation study (qualitative). }We qualitatively observe the
effects of removing individual elements of our optimization on the final mesh surface.}%
\label{fig:ablation}%
\end{figure}

\begin{figure}[t!]
      \centering
      \adjustbox{trim=0cm 0cm 0cm 0cm, clip=true}{%
        \includegraphics[scale=1.1]{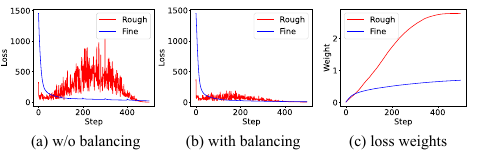}%
        }
      \caption{\textbf{The effect of loss balancing.} We analyze the rough and fine losses throughout optimization without and with our proposed loss balancing in (a) and (b), respectively. (c) shows loss weights values over the course of the optimization in (b).}
      \label{fig:loss_plots}
\end{figure}

We perform an ablation study to assess the impact of our method's different components on the 3D generation process by removing them one at a time and performing mesh generation. These components include the multi-view generated prior images, optimization balancing, rough and fine optimizations, opacity loss, and depth distortion and normal alignment losses. Table \ref{ablation_table} shows our quantitative results on the GSO evaluation set, along with the average number of Gaussians in the learned representations. Fig. \ref{fig:ablation} shows the mesh surfaces for the "train" object generated for each experiment. We analyze the results of each experiment below: \\
\textbf{Multi-View References.} Removing the generated images and running the pipeline only with the input image results in amorphous meshes that lack sharper edges, as shown in Fig. \ref{fig:ablation}c. While the texture and geometry details of the input view are captured relatively well, other unseen views lack detail, causing inconsistencies in quality and content. \\ 
\textbf{Optimization Balancing.} Removing the task-based balancing between rough shape and fine detail optimizations results in less efficient Gaussian representations that contain more Gaussians while missing some details, as shown in Fig. \ref{fig:ablation}d. Fig. \ref{fig:loss_plots} illustrates the behavior of loss weights and how they affect the optimization by comparing loss values. The results show that SDS loss for rough shape optimization is highly unstable. The automatic weight updates lead to more stable optimization, toning down the SDS loss by increasing its weight at a higher rate while also stabilizing local peaks in the fine detail optimization caused by Gaussian densification. \\
\textbf{Rough Shape Optimization.} When the multi-view SDS loss for rough shape optimization is removed, the optimization fails to form to base shape of the object, resulting in noisy, half-formed surfaces in the reconstructed meshes, as visible in Fig. \ref{fig:ablation}e. The lack of a clear base shape also leads to an insufficiently low Gaussian count in the optimized Gaussian representation. \\ 
\textbf{Fine Detail Optimization.} Removing the fine detail optimization, comprised of the multi-view reconstruction and mask losses, results in amorphous meshes that lack sharp texture and geometry details, obtained from representations with large Gaussian counts. The lack of mask loss can also result in floaters, as seen in Fig. \ref{fig:ablation}f. \\ 
\noindent\textbf{Opacity, Depth Distortion and Normal Alignment Losses.} Removing these losses lead to slightly less refined representations with higher Gaussian counts as a result of the lack of alignment. Fig. \ref{fig:ablation}g and \ref{fig:ablation}h show that this affects the smoothness of the meshes, especially on flat surfaces and sharp edges. \\
%\indent Like other methods that rely on multi-view generated images, our method is bound by limitations such as low-resolution outputs and strict requirements on elevation and field-of-view angles of input images.
\vspace{-0.5cm}
\begin{table}[t!]
\normalsize
\caption{Ablation study (quantitative) on the effects of removing individual elements of our optimization.}
\label{ablation_table}
\begin{center}
\resizebox{\columnwidth}{!}{
\begin{tabular}{@{}lcccc@{}}\toprule 
Method & LPIPS $\downarrow$ & CLIP Sim. $\uparrow$ & CLIP Cons. $\uparrow$ & Avg. Gauss. Count \\

\cmidrule{1-5} 
w/o multi-view & 0.15 & 0.86 & 0.82 & 15955\\
w/o opt. balancing & 0.13 & 0.90 & 0.87 & 19757\\
w/o fine opt. & 0.22 & 0.75 & 0.72 & 20900\\
w/o rough opt. & 0.14 & 0.87 & 0.83 & 4280\\
w/o opacity loss & 0.13 & 0.90 & 0.87 & 15968\\
w/o depth loss & 0.13 & 0.90 & 0.87 & 15544\\
\hline
Full loss & \textbf{0.13} & \textbf{0.90} & \textbf{0.87} & \textbf{15489}\\

\bottomrule
\end{tabular}
}

\end{center}
\end{table}

\subsubsection{Multi-View Generation in the Initial Stage}

We perform an ablation study to observe the impact of the multi-view prior images generated in the initial stage on the performance of our method. For this purpose, we repeat our experiment on the GSO dataset using different multi-view generation methods in the initial stage of ConsistentDreamer. We perform experiments with 6 and 4 generated views. To generate 6 views, in addition to our default usage of Zero123++ \cite{shi2023zero123plus}, we also use Wonder3D's \cite{Long_2024_CVPR} multi-view generation model. To perform 4-view generation, we use Zero123++ by discarding two of the generated views, and ImageDream \cite{wang2023imagedream}, the multi-view generation method used in the initial stage of LGM\cite{tang2024lgm}. In addition, we test our method with ground truth images instead of generated images using 4 and 6 views from the GSO dataset. For these experiments, we do not report CLIP consistency as there is no single input image. Our results, along with the used viewpoints for each multi-view generation method are available in Table \ref{quantitive_mv_table}. The results show that our method is highly adaptable to initial multi-view images from various sources and not bound by a specific multi-view generation method, fixed number of views or a fixed set of viewpoints. ConsistentDreamer's performance with images generated by Wonder3D and ImageDream exceed the Wonder3D and LGM results from Table \ref{quantitive_table_gso} respectively, further enforcing our method's capability to transfer multi-view image features from various sources to a 3D mesh. Our experiments on GSO ground truth images show our method can be used for multi-view-to-3D reconstruction when multiple views are available for a significant increase in quality.

\begin{table*}[tb!]
%\normalsize
\caption{Ablation study on multi-view generation method used in the initial stage. ($\beta$: elevation angle, $\theta$: azimuth angle)}
\label{quantitive_mv_table}
\begin{center}
\resizebox{2\columnwidth}{!}{
    \begin{tabular}{@{}lclccccc@{}}\toprule
    Method & \#views & \multicolumn{1}{c}{viewpoints ($\beta$, $\theta$)} & SSIM $\uparrow$ & PSNR $\uparrow$ & LPIPS $\downarrow$ & CLIP Sim. $\uparrow$ & CLIP Cons. $\uparrow$ \\
    \cmidrule{1-8} 
    ImageDream \cite{wang2023imagedream} & 4 & (0$^\circ$, 0$^\circ$), (0$^\circ$, 90$^\circ$), (0$^\circ$, 180$^\circ$), (0$^\circ$, 270$^\circ$) & 0.88 & 18.30 & 0.15 & 0.88 & 0.85 \\
    Zero123++ \cite{shi2023zero123plus} & 4 & (20$^\circ$, 30$^\circ$), ($-$10$^\circ$, 90$^\circ$), (20$^\circ$, 150$^\circ$), (20$^\circ$, 270$^\circ$) & 0.88 & 18.79 & 0.14 & \textbf{0.90} & \textbf{0.87} \\
    GSO \cite{Downs2022GoogleSO} ground truth & 4 & (0$^\circ$, 0$^\circ$), (0$^\circ$, 90$^\circ$), (0$^\circ$, 180$^\circ$), (0$^\circ$, 270$^\circ$) & 0.89 & \textbf{21.51} & \textbf{0.12} & \textbf{0.94} & - \\
    \cmidrule{1-8} 
    Wonder3D\cite{Long_2024_CVPR} & 6 & (0$^\circ$, 0$^\circ$), (0$^\circ$, 45$^\circ$), (0$^\circ$, 90$^\circ$), (0$^\circ$, 180$^\circ$), (0$^\circ$, 270$^\circ$), (0$^\circ$, 315$^\circ$) & \textbf{0.89} & 18.92 & 0.14 & 0.89 & 0.86 \\
    Zero123++\cite{shi2023zero123plus} & 6 & (20$^\circ$, 30$^\circ$), ($-$10$^\circ$, 90$^\circ$), (20$^\circ$, 150$^\circ$), ($-$10$^\circ$, 210$^\circ$), (20$^\circ$, 270$^\circ$), ($-$10$^\circ$, 330$^\circ$) & 0.88 & \textbf{19.06} & \textbf{0.13} & \textbf{0.90} & \textbf{0.87} \\
    GSO \cite{Downs2022GoogleSO} ground truth & 6 & (0$^\circ$, 0$^\circ$), (0$^\circ$, 45$^\circ$), (0$^\circ$, 90$^\circ$), (0$^\circ$, 180$^\circ$), (0$^\circ$, 270$^\circ$), (0$^\circ$, 315$^\circ$) & \textbf{0.90} & 21.35 & \textbf{0.12} & \textbf{0.94} & - \\
    \bottomrule
    \end{tabular}
}
\end{center}
\end{table*}
%\vspace{-0.5cm}
\section{Conclusions}
We introduce ConsistentDreamer, a Gaussian optimization method for 3D asset generation from a single image. By leveraging generated multi-view images for both rough shape and fine detail optimizations, our method outperforms the state-of-the-art in terms of content and quality consistency across views. Dynamic task weights, based on homoscedastic uncertainty, automatically balance these optimizations, resulting in more efficient representations. Our method addresses the challenge of view consistency in 3D asset generation, paving the way for better integration with embodied AI simulations. Moving forward, we aim to explore other 3D representations, such as classical ones like surfels and modern ones like FlexiCubes \cite{shen2023flexicubes} to address the limitations of Gaussian representations and improve compatibility with industrial applications. 

\indent Please refer to the attached supplementary video for additional qualitative results.
\begin{comment}
\section*{CRediT authorship contribution statement}
\textbf{Onat Şahin:} Writing-original draft, Methodology. \textbf{Mohammad Altillawi:} Writing-review and editing, Methodology. \textbf{George Eskandar:} Writing–review and editing. \textbf{Carlos Carbone:} Writing–review and editing.  \textbf{Ziyuan Liu:} Supervision
\section*{Declaration of competing interest}
The authors declare that they have no known competing financial interests or personal relationships that could have appeared to influence the work reported in this paper.
\end{comment}

\section*{Acknowledgments}
The authors of this paper are employed by Intelligent Cloud Technologies Laboratory, Huawei Munich Research Center, part of Huawei Technologies Duesseldorf GmbH.

\begin{comment}

\end{comment}

\bibliographystyle{elsarticle-num}
\bibliography{bibliography}

\end{document}